\documentclass{article}

\usepackage{microtype}
\usepackage{graphicx}
\usepackage{subcaption}
\usepackage{booktabs} 

\usepackage{algorithm}
\usepackage{algorithmic}
\usepackage{amsmath}
\usepackage{amssymb}
\usepackage{multirow}
\usepackage{multicol}

\usepackage{xspace}
\newcommand{\myparagraph}[1]{\vspace{1mm}\noindent\textbf{#1}}
\usepackage{hyperref}

\usepackage[accepted]{icml2026}

\usepackage{amsmath}
\usepackage{amssymb}
\usepackage{mathtools}
\usepackage{amsthm}

\usepackage[capitalize,noabbrev]{cleveref}

\theoremstyle{plain}

\theoremstyle{definition}

\theoremstyle{remark}

\usepackage[textsize=tiny]{todonotes}

\icmltitlerunning{Submission and Formatting Instructions for ICML 2026}

\begin{document}

\twocolumn[
  \icmltitle{Semantically Compatible Knowledge Distillation for Cross-Domain Object Detection with Vision Foundation Models}



  \icmlsetsymbol{equal}{*}

  \begin{icmlauthorlist}
    \icmlauthor{Qifeng Zhang}{sch}
    \icmlauthor{Ting Xiang}{sch}
    \icmlauthor{Zeyuan Bai}{sch}
    \icmlauthor{Changjian Chen}{sch}
  \end{icmlauthorlist}

  \icmlaffiliation{sch}{National Supercomputing Center in Changsha, Hunan University, Changsha, Hunan Province, China}

  \icmlcorrespondingauthor{Changjian Chen}{changjianchen@hnu.edu.cn}
  \icmlcorrespondingauthor{Qifeng Zhang}{zqfhnucsee@hnu.edu.cn}

  \icmlkeywords{Machine Learning, ICML}

  \vskip 0.3in
]



\printAffiliationsAndNotice{}  

\begin{abstract}
Vision foundation models (VFMs) offer strong generalization capabilities for domain-adaptive object detection (DAOD).
However, existing VFM-based methods overlook the spatial-scale discrepancy between teacher and student feature maps, resulting in semantic incompatibility that weakens both feature alignment and pseudo-label learning.
Moreover, domain shift can cause source-trained VFM teachers to miss target-domain objects, limiting the quality of their pseudo-labels.
To address these issues, we propose the Semantic Localization-Enhanced Teacher (SLE-T), a semantically compatible knowledge-distillation framework built around a lightweight SLE Adapter for DINOv2.
SLE Adapter injects pretrained local-texture priors into DINOv2 to improve cross-domain recognition and reformulates its features into dense representations that are spatially and semantically compatible with the student detector.
SLE-T transfers the resulting teacher knowledge through either pseudo-label learning or feature alignment.
We instantiate SLE-T with DINOv2-B and DINOv2-L (the ViT-B and ViT-L variants) and compare them with the larger DINOv2-G teacher.
Extensive experiments on three DAOD benchmarks demonstrate that our method achieves state-of-the-art performance, and ablation studies confirm the importance of teacher--student semantic compatibility.
Notably, SLE-T with DINOv2-B produces competitive or superior pseudo-labels using approximately one-quarter of the training time of DINOv2-G and substantially less GPU memory, demonstrating efficient VFM knowledge transfer under limited computational resources.
Our code is available at \url{https://github.com/hnu-vis/SLE-T}.
\end{abstract}

\begin{figure*}[t!]  
    \centering
    \includegraphics[width=1.0\textwidth]{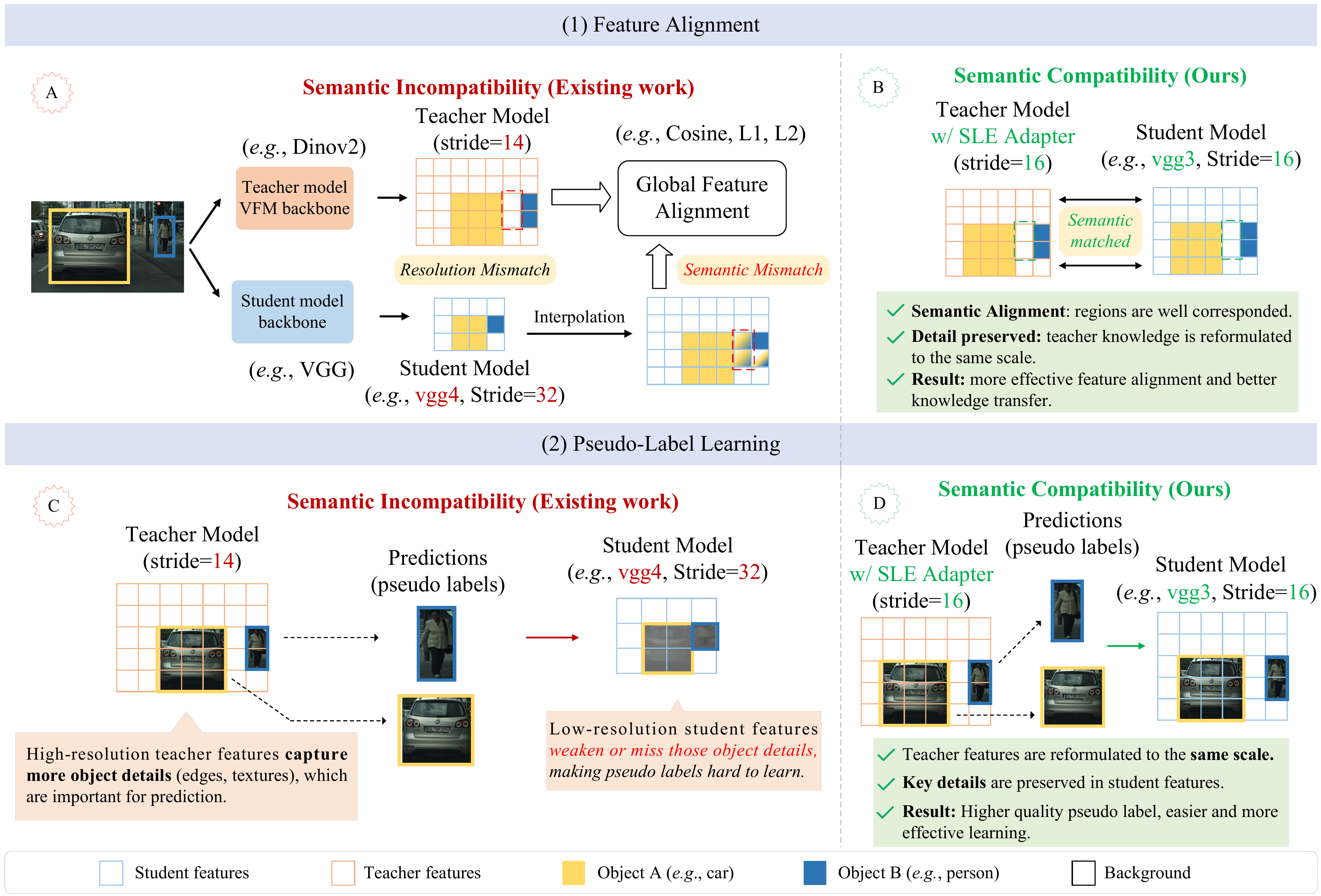}
    \caption{Motivation for semantically compatible knowledge transfer. (A) Resizing matches feature dimensions but not object-level semantics. (B) Our method reformulates teacher features and selects a stride-compatible student representation. (C) Coarse student features lose object details which are helpful for pseudo-label learning. (D) Compatible representations preserve these details for more effective transfer.}
    \label{fig:intro}
\end{figure*}

\section{Introduction}

Domain adaptation object detection (DAOD) transfers a detector from a labeled source domain to an unlabeled target domain, enabling deployment in changing environments without costly target annotations.
Vision foundation models (VFMs), particularly DINOv2~\cite{oquab2023dinov2}, provide transferable representations and are increasingly used as DAOD teachers~\cite{lavoie2025large, cui2026expert}.
However, transferable teacher features alone do not guarantee effective knowledge transfer; they must also be compatible with the spatial granularity and task semantics of the student detector~\cite{hao2023one, yang2022focal, shu2021channel}.

Existing VFM-based DAOD methods mainly transfer knowledge through feature alignment and pseudo-label learning~\cite{lavoie2025large, cui2026expert}.
%
We identify an overlooked spatial-scale discrepancy in both mechanisms due to the different resolutions and strides.
Specifically, for feature alignment, Interpolation can match their tensor dimensions, but cannot establish correspondence between teacher features and student feature locations (Fig.~\ref{fig:intro}A). 
The resulting semantic incompatibility forces features with different spatial granularity and task semantics to align, weakening feature-level knowledge transfer.
For pseudo-label learning, the student must assimilate object-level supervision generated from an incompatible teacher representation, limiting performance (Fig.~\ref{fig:intro}C). 
Our controlled experiments confirm that improving teacher–student semantic compatibility benefits both feature alignment and pseudo-label learning.

Our central insight is that \textbf{VFM features should be reformulated into representations that are spatially and semantically compatible with the student, rather than merely resized to the same dimensions.}
This requires preserving transferable VFM semantics, adapting the teacher's spatial granularity, and improving target-domain recognition without the cost of fully adapting a large VFM.
To this end, we propose the Semantic Localization-Enhanced Teacher (SLE-T), a framework comprising SLE Adapter and two alternative knowledge-transfer routes.
SLE Adapter injects pretrained local-texture priors into DINOv2 to reduce missed target-domain objects and extracts dense stride-16 features that match the student detector's spatial granularity and task semantics.
The resulting teacher generates higher-quality pseudo-labels and provides semantically compatible targets for feature alignment, while most VFM parameters remain frozen for efficient training.
We instantiate SLE-T with DINOv2-B and DINOv2-L and compare them with the larger DINOv2-G teacher.

Extensive experiments demonstrate that SLE-T improves both the effectiveness and efficiency of VFM-based DAOD. 
Our method achieves state-of-the-art performance on three DAOD benchmarks, and the ablation studies verify that semantic compatibility improves both feature alignment and pseudo-label learning.
Moreover, SLE-T with DINOv2-B provides competitive or superior pseudo-labels with approximately one-quarter of the training time of DINOv2-G and substantially lower memory use.

Our contributions are summarized as follows:

1. We identify teacher--student semantic incompatibility as a key limitation of VFM-based DAOD and show that it weakens both feature alignment and pseudo-label learning.

2. We propose SLE-T, which combines SLE Adapter with pseudo-label learning or feature alignment for semantically compatible knowledge transfer.

3. We achieve state-of-the-art performance on three DAOD benchmarks and show that an SLE-enhanced DINOv2-B teacher can match or surpass the larger DINOv2-G teacher with substantially lower computational cost.

\begin{figure*}[t!]
    \centering
    \includegraphics[width=1.0\textwidth]{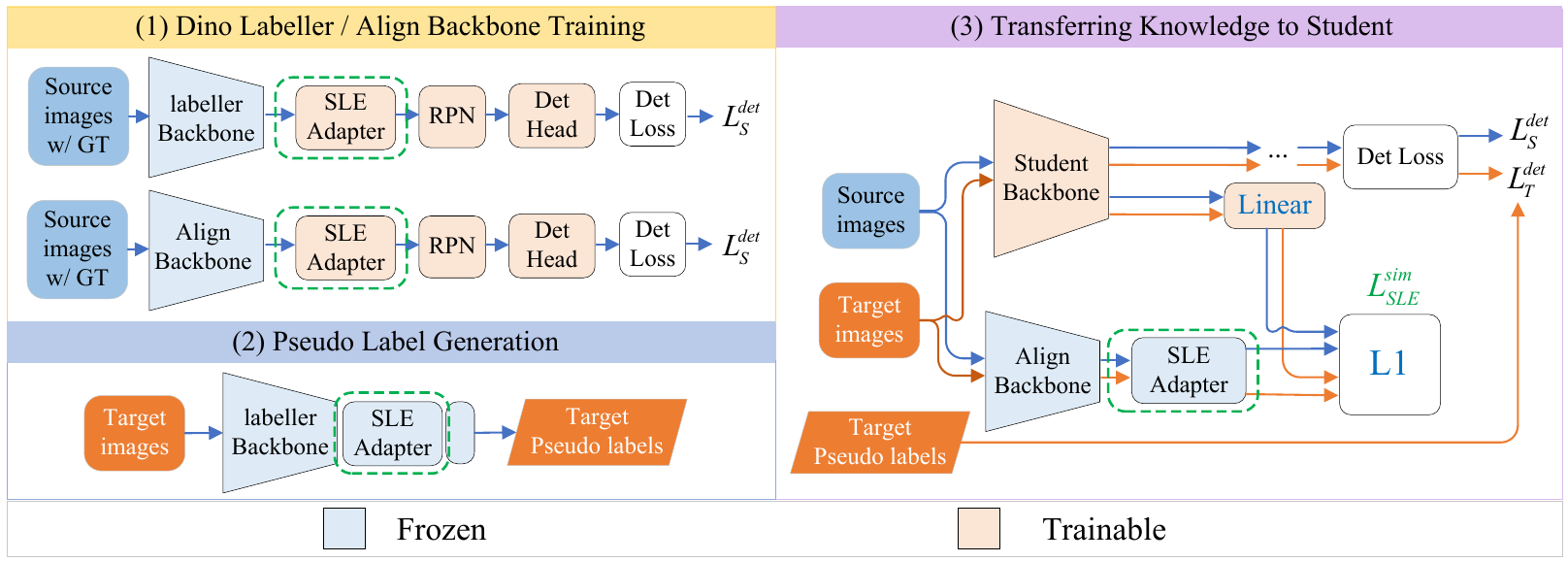}
    \caption{Overview of the SLE Teacher framework. An SLE-enhanced teacher is trained on source annotations and then frozen to transfer knowledge through either pseudo-label learning or feature alignment.}
    \label{fig:training_pipeline}
\end{figure*}

\section{Related Work}

\subsection{Domain-Adaptive Object Detection}
DAOD methods primarily follow two directions: domain-discrepancy reduction and self-training.
Domain-discrepancy methods learn domain-invariant representations through adversarial learning~\cite{chen2025datr, he2025differential, zhou2022multi}, contrastive learning~\cite{cao2023contrastive, bai2025contrastive}, or graph alignment~\cite{li2022scan, gao2023csda}.
Self-training methods use a teacher to generate target-domain pseudo-labels for a student detector~\cite{chen2025refining, deng2023harmonious, kennerley2024cat, li2022cross}.
Recent approaches improve this process through pseudo-label reweighting~\cite{deng2023harmonious}, confidence alignment~\cite{chen2025refining}, and class-wise balancing~\cite{kennerley2024cat}.
Unlike these methods, our work studies how a VFM teacher should represent and transfer knowledge to a detection student.

\subsection{Vision Foundation Models for DAOD}
Prior DAOD methods obtain transferable supervision from vision-language models~\cite{fahes2023poda, vidit2023clip, yang2024unified, zhang2025upre, sikdar2025picazo} or diffusion models~\cite{he2024diffusion, he2025generalized, he2025boosting}.
Our work instead focuses on DINOv2, a self-supervised VFM with strong cross-domain representations~\cite{oquab2023dinov2}.

DINO Teacher (DT)~\cite{lavoie2025large} uses DINOv2 to generate target pseudo-labels and supervise student features, whereas ETS~\cite{cui2026expert} combines VFM knowledge with an EMA teacher to provide domain-specific guidance.
These methods improve the source of supervision but do not explicitly address the spatial-scale discrepancy between VFM teachers and detection students.
Matching feature dimensions by interpolation leaves this semantic incompatibility unresolved, limiting both feature alignment and the student's assimilation of pseudo-label supervision.
Moreover, a VFM teacher trained only on the source domain can miss target-domain objects and consequently produce incomplete pseudo-labels.

In this work, we propose SLE-T, which uses the proposed SLE Adapter to inject local-texture priors into DINOv2 and extract dense representations compatible with the student detector.
This design positions our method as a teacher-representation adaptation approach that improves both feature alignment and pseudo-label quality while reducing the cost of VFM-based DAOD.

\section{Method}

\subsection{Problem Setting}
Let $D_s=\{(X_s^i,B_s^i,Y_s^i)\}_{i=1}^{n_s}$ denote a labeled source domain and $D_t=\{X_t^i\}_{i=1}^{n_t}$ an unlabeled target domain, where $X$, $B$, and $Y$ denote images, bounding boxes, and class labels, respectively.
Our goal is to learn a detector $f_\theta$ that predicts target-domain boxes and labels, $f_\theta(X_t)=(\hat B_t,\hat Y_t)$.
Effective transfer requires compatible teacher and student representations.
On the teacher side, we reformulate DINOv2 features only from stride 14 to stride 16, limiting spatial distortion while matching the detector's resolution.
On the student side, we use the stride-16 feature whose resolution is closest to the DINOv2 output, rather than a coarser detection feature.
We further attach lightweight trainable modules to DINOv2-B or DINOv2-L, avoiding the cost of training a DINOv2-G labeller while improving target-domain recognition.
SLE-T comprises source-domain training of an SLE-enhanced teacher and knowledge transfer to $f_\theta$ through either pseudo-label learning or feature alignment.
%
Figure~\ref{fig:training_pipeline} summarizes the workflow.
First, the SLE-enhanced labeller or alignment teacher is trained with source annotations; the frozen labeller then generates target pseudo-labels; finally, the student receives either output-level pseudo-label supervision or feature-level alignment supervision.

\begin{figure*}[t!]  
    \centering
    \includegraphics[width=1.0\textwidth]{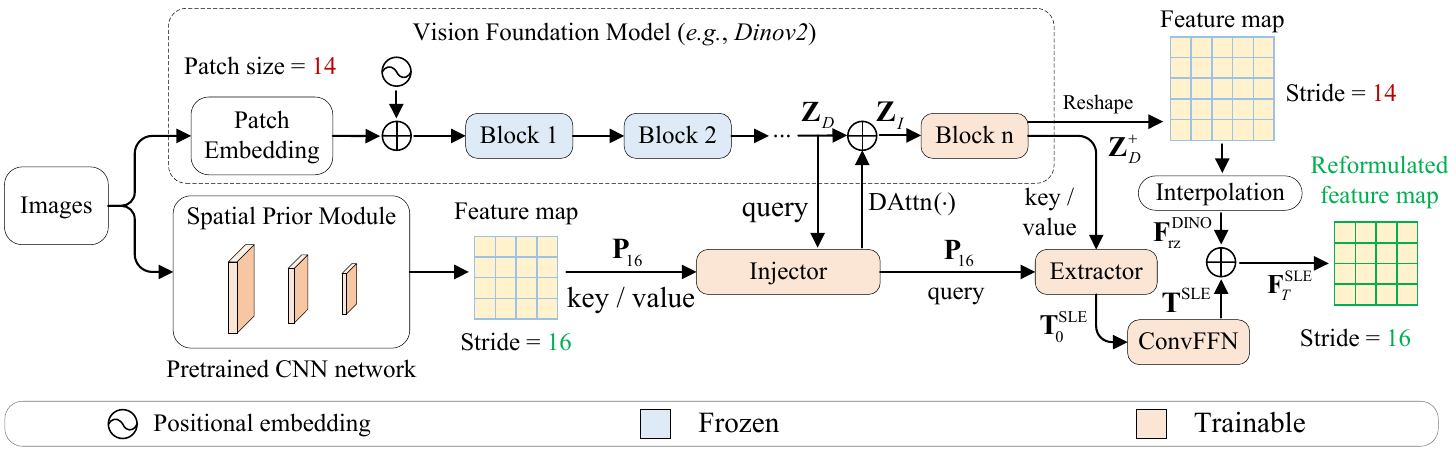}
    \caption{Architecture of SLE Adapter. The Spatial Prior Module extracts stride-16 local features from a pretrained CNN. The Injector uses deformable attention ($\mathrm{DAttn}$) to inject these local cues into DINOv2 tokens, while the Extractor reverses the attention direction to produce a dense stride-16 representation refined by ConvFFN. The resized final DINOv2 feature is added as a semantic residual, preserving pretrained VFM knowledge in the student-compatible teacher feature. Blue and orange blocks denote frozen and trainable components, respectively.}
    \label{fig:sle}
\end{figure*}

\subsection{SLE Adapter}

Raw DINOv2 tokens and dense student features differ in spatial granularity and task semantics, creating semantic incompatibility.
SLE Adapter addresses this problem with a Spatial Prior Module (SPM), an Injector, and an Extractor (Fig.~\ref{fig:sle}).
Together, they inject local texture cues into DINOv2 and minimally reformulate its stride-14 tokens into a semantically compatible stride-16 teacher representation.

\myparagraph{Spatial Prior Module.}
For an image batch $X\in\mathbb{R}^{B\times3\times H\times W}$, a pretrained VGG pathway extracts local convolutional features.
We select its stride-16 feature because its spatial granularity is close to the stride-14 DINOv2 tokens and the stride-16 student feature.
A $1\times1$ convolution projects it to the DINOv2 channel dimension $C_D$, after which it is flattened into local-prior tokens:
\begin{equation}
\begin{aligned}
    \mathbf{P}_{16}
    &=
    \mathrm{Flatten}\!\left(
    \mathrm{Conv}_{1\times1}(E_{\mathrm{VGG}}^{(16)}(X))
    \right)
    \in
    \mathbb{R}^{B \times N_{16} \times C_D},
    \\
    &\qquad N_{16}=H_{16}W_{16}.
\end{aligned}
\end{equation}
where $H_{16}=H/16$ and $W_{16}=W/16$.

\myparagraph{Local-Texture Injection.}
Let $\mathbf{Z}_D\in\mathbb{R}^{B\times N_D\times C_D}$ denote the tokens produced by the frozen DINOv2 blocks before the final block.
The Injector uses deformable attention, denoted by $\mathrm{DAttn}$, to inject the stride-16 VGG3 prior $\mathbf P_{16}$ into these tokens.
For each DINOv2 query, $\mathrm{DAttn}$ predicts sampling offsets and attention weights, samples a small set of relevant locations from $\mathbf P_{16}$, and aggregates their local texture cues.
Only the VGG3 feature level is used; no multi-scale VGG features participate.
\begin{equation}
    \mathbf{Z}_I
    =
    \mathbf{Z}_D
    +
    \gamma_I
    \mathrm{DAttn}
    \left(
    \mathrm{LN}(\mathbf{Z}_D),
    \mathbf{r}_{D\to16},
    \mathrm{LN}(\mathbf{P}_{16}),
    \Omega_{16}
    \right),
\end{equation}
where $\gamma_I$ is a learnable residual scale, $\mathrm{LN}$ is Layer Normalization operation, $\mathbf r_{D\to16}$ contains reference points from the DINOv2 grid to the stride-16 grid, and $\Omega_{16}$ describes that grid.
The trainable final DINOv2 block $\mathcal B_D^{L-1}$ then produces the enhanced tokens
\begin{equation}
    \mathbf{Z}_D^+
    =
    \mathrm{LN}_D\left(\mathcal{B}_D^{L-1}(\mathbf{Z}_I)\right).
\end{equation}

\myparagraph{Student-Compatible Feature Extraction.}
The Extractor reverses the attention direction: $\mathbf P_{16}$ supplies the queries and $\mathbf Z_D^+$ supplies the values.
Here, $\mathrm{DAttn}$ lets each VGG3 query sample relevant locations from the enhanced DINOv2 tokens and aggregate their semantic information.
Because the queries remain on the single stride-16 VGG3 grid, the output preserves the student-compatible spatial layout:
\begin{equation}
    \mathbf{T}^{\mathrm{SLE}}_0
    =
    \mathbf{P}_{16}
    +
    \mathrm{DAttn}
    \left(
    \mathrm{LN}(\mathbf{P}_{16}),
    \mathbf{r}_{16\to D},
    \mathrm{LN}(\mathbf{Z}_D^+),
    \Omega_D
    \right),
\end{equation}
where $\mathbf r_{16\to D}$ and $\Omega_D$ specify the DINOv2 token grid.
A convolutional feed-forward residual refines the extracted tokens:
\begin{equation}
    \mathbf{T}^{\mathrm{SLE}}
    =
    \mathbf{T}^{\mathrm{SLE}}_0
    +
    \mathrm{ConvFFN}
    \left(
    \mathrm{LN}(\mathbf{T}^{\mathrm{SLE}}_0),
    H_{16},
    W_{16}
    \right).
\end{equation}
We reshape the Extractor output and interpolate the final stride-14 DINOv2 feature to the same stride-16 resolution:
\begin{equation}
    \begin{aligned}
    \mathbf F_{\mathrm{Ext}}^{\mathrm{SLE}}
    &=
    \mathrm{Reshape}(\mathbf T^{\mathrm{SLE}}),\\
    \mathbf F_{\mathrm{rz}}^{\mathrm{DINO}}
    &=
    \mathrm{Interp}\!\left(
    \mathrm{Reshape}(\mathbf Z_D^+),H_{16},W_{16}
    \right).
    \end{aligned}
\end{equation}
Their residual fusion forms the final teacher feature:
\begin{equation}
    \mathbf F_T^{\mathrm{SLE}}
    =
    \mathbf F_{\mathrm{Ext}}^{\mathrm{SLE}}
    +
    \mathbf F_{\mathrm{rz}}^{\mathrm{DINO}}
    \in
    \mathbb R^{B\times C_D\times H_{16}\times W_{16}}.
\end{equation}
The resized DINOv2 feature acts as a semantic anchor that preserves pretrained VFM knowledge, while the Extractor supplies local, detection-oriented refinements.

\myparagraph{Student-Side Feature Selection.}
The teacher-side reformulation is paired with the stride-16 feature of the student, which is spatially closer to DINOv2's stride-14 output than the stride-32 feature.
This choice reduces the spatial transformation required for knowledge transfer: the stride-16 feature serves as the alignment interface in the feature-alignment route and as the detector representation trained by pseudo-label supervision in the pseudo-label route.
The similar strides facilitate semantic compatibility but do not assume that teacher and student features are semantically identical.
\begin{table*}[htbp]
    \centering
    \setlength{\tabcolsep}{4.5pt}
    \renewcommand{\arraystretch}{1.1}

    \begin{tabular}{llccccccccc}
        \toprule
        \textbf{Method}
        & \textbf{Type}
        & \textbf{Person}
        & \textbf{Rider}
        & \textbf{Car}
        & \textbf{Truck}
        & \textbf{Bus}
        & \textbf{Train}
        & \textbf{Motor}
        & \textbf{Bicycle}
        & \textbf{mAP} \\
        \midrule

        DA-Faster~\cite{chen2018domain}
        & FR
        & 29.2 & 40.4 & 43.4 & 19.7 & 38.3 & 28.5 & 23.7 & 32.7 & 32.0 \\

        DICN~\cite{jiao2022dual}
        & FR
        & 47.3 & 57.4 & 64.0 & 22.7 & 45.6 & 29.6 & 38.6 & 47.4 & 44.1 \\

        NLTE~\cite{liu2022towards}
        & FR+GA
        & 43.1 & 50.7 & 58.7 & 33.6 & 56.7 & 42.7 & 33.7 & 43.3 & 45.4 \\

        PT~\cite{chen2022learning}
        & FR+MT
        & 43.2 & 52.4 & 63.4 & 33.4 & 56.6 & 37.8 & 41.3 & 48.7 & 47.1 \\

        MIC~\cite{hoyer2023mic}
        & FR+MT
        & 50.9 & 55.3 & 67.0 & 33.9 & 52.4 & 33.7 & 40.6 & 47.5 & 47.6 \\

        AT~\cite{li2022cross}
        & FR+MT
        & 45.5 & 55.1 & 64.2 & 35.0 & 56.3 & 54.3 & 38.5 & 51.9 & 50.9 \\

        CMT~\cite{cao2023contrastive}
        & FR+MT
        & 47.0 & 55.7 & 64.5 & 39.4 & 63.2 & 51.9 & 40.3 & 53.1 & 51.9 \\

        REACT~\cite{li2024react}
        & FR+MT
        & 51.4 & 57.9 & 67.4 & 37.7 & 58.4 & 52.8 & 44.6 & 54.6 & 53.1 \\

        DT$^{*}$~\cite{lavoie2025large}
        & FR+MT (ViT-G)
        & 46.5
        & 56.2
        & 62.3
        & 40.8
        & 64.1
        & \textbf{58.0}
        & 45.5
        & 51.8
        & 53.2 \\

        ETS~\cite{cui2026expert}   
        & FR +MT 
        & 49.0  
        & 61.8  
        & 65.8  
        & 41.7  
        & \textbf{69.9} 
        & 56.8  
        & 46.7  
        & 57.7  
        & 56.2 \\

        \midrule

        SLE-T (Ours)
        & FR+MT (ViT-B)
        & \textbf{57.0}
        & \textbf{65.1}
        & \textbf{70.0}
        & 42.9
        & 63.2
        & 52.9
        & 48.5
        & \textbf{60.0}
        & 57.4 \\

        SLE-T (Ours)
        & FR+MT (ViT-L)
        & 56.7
        & 64.5
        & 69.8
        & \textbf{43.6}
        & 62.3
        & 53.7
        & \textbf{51.1}
        & 59.3
        & \textbf{57.6} \\

        \bottomrule
    \end{tabular}%
    \caption{Class-wise $\mathrm{AP}_{50}$ (\%) on Cityscapes $\rightarrow$ Foggy Cityscapes (all fog levels). GA and MT denote Graph Alignment and Mean Teacher framework, respectively. `*' indicates results we reproduced using the official publicly available code. }
    \label{tab:c2fc}
\end{table*}

\subsection{SLE Teacher Knowledge Transfer}
\textbf{Teacher Training.}
Training a large DINOv2-G labeller is expensive, yet a source-trained labeller may still miss target objects.
As shown in stage (1) of Fig.~\ref{fig:training_pipeline}, we instead construct $g_{\mathrm{SLE}}$ from DINOv2-B or DINOv2-L, SLE Adapter, and a detection head, and train it on source annotations:
\begin{equation}
    \mathcal L_{\mathrm{teacher}}
    =
    \mathcal L_S^{\mathrm{det}}(g_{\mathrm{SLE}}(X_s),B_s,Y_s).
\end{equation}
All DINOv2 blocks except the final block remain frozen, while the final block, SPM, Injector, Extractor, and detection head are optimized.
This parameter-efficient design shortens labeller training, while the injected local texture cues improve its sensitivity to target-domain objects.
The trained teacher is then frozen and used in one of the following transfer routes.

\myparagraph{Pseudo-Label Learning.}
In stage (2) of Fig.~\ref{fig:training_pipeline}, the frozen teacher predicts target pseudo-labels
\begin{equation}
    (\tilde B_t,\tilde Y_t)=g_{\mathrm{SLE}}(X_t).
\end{equation}
Because SLE Adapter combines DINOv2 semantics with local texture cues, these predictions reduce missed target objects and provide stronger localization-sensitive supervision than labels generated from raw DINOv2 features.
The student is trained with source annotations and target pseudo-labels:
\begin{equation}
    \mathcal L_{\mathrm{PL}}
    =
    \mathcal L_S^{\mathrm{det}}
    +
    \lambda^{\mathrm{unsup}}\mathcal L_T^{\mathrm{det}},
\end{equation}
where $\mathcal L_T^{\mathrm{det}}$ uses $(\tilde B_t,\tilde Y_t)$ as target supervision.

\myparagraph{Feature Alignment.}
In the feature-alignment route shown in stage (3) of Fig.~\ref{fig:training_pipeline}, the frozen teacher provides dense features $\mathbf F_T^{\mathrm{SLE}}(X)$ rather than pseudo-labels.
Let $\mathbf F_S(X)$ be the stride-16 student feature.
A shape-preserving linear projection $\phi_{\mathrm{lin}}$ maps its channels to $C_D$, yielding the same shape as the teacher feature:
\begin{equation}
    \phi_{\mathrm{lin}}(\mathbf F_S(X)),
    \mathbf F_T^{\mathrm{SLE}}(X)
    \in
    \mathbb R^{B\times C_D\times H_{16}\times W_{16}}.
\end{equation}
We minimize their L1 distance:
\begin{equation}
    \mathcal{L}^{sim}_{\mathrm{SLE}}(X)
    =
    \left\|
    \phi_{\mathrm{lin}}(\mathbf{F}_S(X))
    -
    \mathrm{sg}\left(\mathbf{F}_T^{\mathrm{SLE}}(X)\right)
    \right\|_1,
\end{equation}
where $\mathrm{sg}$ stops gradients through the teacher.
Following DINO Teacher, alignment is applied to source images initially and to both source and target images in the later stage.
For an alignment batch $\mathcal B_A$, where $\mathcal B_A=\mathcal B_S$ initially and $\mathcal B_A=\mathcal B_S\cup\mathcal B_T$ later, the student objective is
\begin{equation}
    \mathcal L_{\mathrm{FA}}
    =
    \mathcal L_S^{\mathrm{det}}
    +
    \frac{\lambda^{\mathrm{sim}}}{|\mathcal B_A|}
    \sum_{X\in\mathcal B_A}
    \mathcal L_{\mathrm{SLE}}^{sim}(X).
\end{equation}
Thus, $\mathcal L_{\mathrm{PL}}$ transfers output-level predictions, whereas $\mathcal L_{\mathrm{FA}}$ transfers intermediate representations.
We treat them as independent routes: FA directly evaluates representation-level transferability, while PL is adopted for the final configuration because it yields the best detection performance.

At inference, the SLE-enhanced teacher and projection layer are discarded, and only the adapted student detector $f_\theta$ is retained.
The complete SLE-T training procedure is summarized in the supplemental material.

\section{Experiments}

\subsection{Datasets}
\myparagraph{Cityscapes.}
We use Cityscapes as the labeled source domain, with 2,975 training images, 500 validation images, and eight detection categories.

\myparagraph{Foggy Cityscapes.}
Foggy Cityscapes adds three fog levels to Cityscapes and contains 8,925 training and 1,500 validation images in the full setting~(Lavoie et al. 2025).

\myparagraph{BDD100K Daytime.}
Following prior DAOD protocols~\cite{deng2023harmonious,lavoie2025large}, we use 36,728 training and 5,258 validation images from BDD100K Daytime, with seven categories shared with Cityscapes.

\myparagraph{ACDC.}
ACDC contains real fog, night, rain, and snow scenes in the Cityscapes category space; each condition has 400 training images and 100 validation images, except night with 106 validation images~(Lavoie et al. 2025).


\subsection{Experimental Setup}
We implement all models in Detectron2 and follow DINO Teacher (DT)~\cite{lavoie2025large} unless stated otherwise.
Faster R-CNN uses ImageNet-pretrained VGG16 for Foggy Cityscapes and BDD100K Daytime and ResNet-50 for ACDC; we select the stride-16 VGG3 or Res3 feature as the student representation.
Training uses two RTX 4090 GPUs and a batch of four images, equally split between source and target domains.
Images are resized to a height of 560 pixels to accommodate the stride-14 DINOv2 and stride-16 SLE grids.
We train SLE Adapter for 30k iterations with AdamW, a learning rate of $10^{-4}$, weight decay of $0.1$, and a 5k-iteration linear warm-up.
Only the final DINOv2 Transformer block, SLE Adapter, RPN, and RoI head are optimized; the remaining DINOv2 blocks are frozen.
We reimplement DT from its official codebase on the same platform for a controlled comparison.

\subsection{Experimental Results}
We evaluate Cityscapes $\rightarrow$ Foggy Cityscapes / BDD100K Daytime, and Cityscapes $\rightarrow$ ACDC using $\mathrm{mAP}_{50}$.
SLE-T is instantiated with DINOv2-B and DINOv2-L and compared with DINOv2-G.
Unless otherwise stated, Tabs.~\ref{tab:c2fc}, \ref{tab:c2bd}, and~\ref{tab:c2acdc} report the PL-only configuration, which outperforms both FA alone and joint FA--PL training; the corresponding ablation is provided in the supplemental material.

\begin{table*}[htbp]
    \centering
    \setlength{\tabcolsep}{4.5pt}
    \renewcommand{\arraystretch}{1.1}

    \begin{tabular}{llcccccccc}
        \toprule
        \textbf{Method}
        & \textbf{Type}
        & \textbf{Person}
        & \textbf{Rider}
        & \textbf{Car}
        & \textbf{Truck}
        & \textbf{Bus}
        & \textbf{Motor}
        & \textbf{Bicycle}
        & \textbf{mAP} \\
        \midrule

        DA-Faster~\cite{chen2018domain}
        & FR
        & 28.9 & 27.4 & 44.2 & 19.1 & 18.0 & 14.2 & 22.4 & 24.9 \\

        SIGMA~\cite{li2022sigma}
        & FCOS+GA
        & 46.9 & 29.6 & 64.1 & 20.2 & 23.6 & 17.9 & 26.3 & 32.7 \\

        TDD~\cite{he2022cross}
        & FR+MT
        & 39.6 & 38.9 & 53.9 & 24.1 & 25.5 & 24.5 & 28.8 & 33.6 \\

        PT~\cite{chen2022learning}
        & FR+MT
        & 40.5 & 39.9 & 52.7 & 25.8 & 33.8 & 23.0 & 28.8 & 34.9 \\

        NSA~\cite{zhou2023unsupervised}
        & FR+MT
        & -- & -- & -- & -- & -- & -- & -- & 35.5 \\

        REACT~\cite{li2024react}
        & FR+MT
        & -- & -- & -- & -- & -- & -- & -- & 35.8 \\

        CAT~\cite{kennerley2024cat}
        & FR+MT
        & 44.6 & 41.5 & 61.2 & 31.4 & 34.6 & 24.4 & 31.7 & 38.5 \\

        HT~\cite{deng2023harmonious}
        & FCOS+MT
        & 53.4 & 40.4 & 63.5 & 27.4 & 30.6 & 28.2 & 38.0 & 40.2 \\

        DT$^{*}$~\cite{lavoie2025large}
        & FR+MT (ViT-G)
        & 48.5  & 47.2  & 63.9  & 39.7  & 42.9  & 33.1  & 41.0  & 45.2 \\

        \midrule

        SLE-T (Ours)
        & FR+MT (ViT-B)
        & 57.2 & 46.5 & 71.0 & 39.3 & 41.7
        & \textbf{37.8} & 40.2 & 47.7 \\

        SLE-T (Ours)
        & FR+MT (ViT-L)
        & \textbf{57.5}
        & \textbf{47.2}
        & \textbf{71.1}
        & \textbf{40.7}
        & \textbf{44.6}
        & 36.9
        & \textbf{41.9}
        & \textbf{48.6} \\

        \bottomrule
    \end{tabular}%
    \caption{Class-wise $\mathrm{AP}_{50}$ (\%) on Cityscapes $\rightarrow$ BDD100K Daytime.}
    \label{tab:c2bd}
\end{table*}



\begin{table}[htbp]
    \centering
    \begin{tabular}{lcccc}
        \toprule
        \textbf{Method}
        & \textbf{Fog}
        & \textbf{Night}
        & \textbf{Rain}
        & \textbf{Snow} \\
        \midrule
        AT~\cite{li2022cross}
        & 62.2
        & 29.5
        & 37.7
        & 55.2 \\

        DT*~\cite{lavoie2025large}
        & 62.0
        & 33.3
        & 33.4
        & 49.4 \\

        SLE-T (ViT-B)
        & \textbf{66.3}
        & 34.9
        & 38.8
        & 55.5 \\

        SLE-T (ViT-L)
        & 66.0
        & \textbf{35.2}
        & \textbf{42.9}
        & \textbf{55.7} \\
        \bottomrule
    \end{tabular}
    \caption{$\mathrm{mAP}_{50}$ (\%) comparison on Cityscapes $\rightarrow$ ACDC
    under different adverse weather conditions.}
    \label{tab:c2acdc}
\end{table}

\begin{table}[htbp]
    \centering
    \begin{tabular}{lccc}
        \toprule
        \textbf{Backbone}
        & \multicolumn{2}{c}{\textbf{$\mathrm{mAP}_{50}$}}
        & \multirow{2}{*}{\textbf{Training Time (h)}} \\
        \cmidrule(lr){2-3}
        & \textbf{C}
        & \textbf{BD}
        & \\
        \midrule
        ViT-G
        & 62.8
        & 50.7
        & 26.35 \\
        
        SLE (ViT-B)
        & \textbf{65.8}
        & 49.9
        & \textbf{5.93} \\
        
        SLE (ViT-L)
        & 64.9
        & \textbf{52.6}
        & 7.95 \\
        \bottomrule
    \end{tabular}
    \caption{Labeller accuracy and training time on Cityscapes $\rightarrow$ BDD100K Daytime. C and BD denote Cityscapes and BDD100K Daytime.}
    \label{tab:c2bd_tt}
\end{table}

\myparagraph{Cross-Domain Detection.}
Tables~\ref{tab:c2fc}, \ref{tab:c2bd}, and~\ref{tab:c2acdc} show that SLE-T achieves state-of-the-art performance across synthetic fog, cross-camera daytime scenes, and real adverse conditions.
With DINOv2-L, SLE-T reaches 57.6 and 48.6 $\mathrm{mAP}_{50}$ on Foggy Cityscapes and BDD100K Daytime, exceeding DT with DINOv2-G by 4.4 and 3.3, respectively.
On ACDC, both SLE-T variants consistently outperform DT across all four adverse conditions.
The ViT-L variant improves DT by 4.0, 1.9, 9.5, and 6.3 on fog, night, rain, and snow, respectively.
It exceeds ViT-B by 4.1 on rain, while the two variants remain within 0.3 under the other conditions, suggesting that the larger backbone is particularly beneficial for the more complex rain-domain shift.
The supplemental material provides a class-wise ACDC breakdown and qualitative detections under dense fog.

\myparagraph{Efficient Labellers.}
Tab.~\ref{tab:c2bd_tt} shows that SLE with DINOv2-B uses 22.5\% of the DINOv2-G training time while remaining within 0.8 on BDD100K Daytime.
SLE with DINOv2-L uses 30.2\% of the training time and improves target performance by 1.9.
These results support our claim that lightweight adaptation of a smaller VFM can produce stronger target supervision at substantially lower cost.
The supplemental material further visualizes the accuracy--size trade-off and evaluates an SLE-T instantiation with DINOv3-B.

\subsection{Ablation Study}
We design the ablations to test teacher--student semantic compatibility, student feature selection, and the effectiveness of SLE Adapter.
%


\begin{table}[htbp]
    \centering
    \begin{tabular}{cccccc}
        \toprule
        \textbf{No.}
        & \textbf{Layer}
        & \textbf{FA}
        & \textbf{AlignM}
        & \textbf{FC}
        & \textbf{BD} \\
        \midrule
        1 & VGG4 & --           & --                  & 40.1          & 22.3          \\
        2 & VGG4 & $\checkmark$ & ViT-B               & 46.6          & 32.2          \\
        3 & VGG4 & $\checkmark$ & SLE (ViT-B) & 47.4          & 33.1          \\
        \midrule
        4 & VGG3 & --           & --                  & 38.7          & 21.7          \\
        5 & VGG3 & $\checkmark$ & ViT-B               & 48.9          & 33.3          \\
        6 & VGG3 & $\checkmark$ & SLE (ViT-B) & \textbf{51.0} & \textbf{34.3} \\
        \bottomrule
    \end{tabular}
    \caption{Feature-alignment ablation on Foggy Cityscapes (FC) and BDD100K Daytime (BD). FA and AlignM denote feature alignment and the alignment model, respectively.}
    \label{tab:alignment_method_comparison}
\end{table}

\myparagraph{Semantic Compatibility in Feature Alignment.}
Tab.~\ref{tab:alignment_method_comparison} compares raw and SLE-enhanced DINOv2-B as alignment teachers.
Although VGG3 is weaker than VGG4 without alignment, using its stride-16 feature improves raw-DINOv2 alignment by 2.3 on FC and 1.1 on BD.
This reversal indicates that the student feature closest to DINOv2's stride-14 resolution provides a more compatible interface for knowledge transfer.

\begin{figure}
    \centering
    \includegraphics[width=0.48\textwidth]{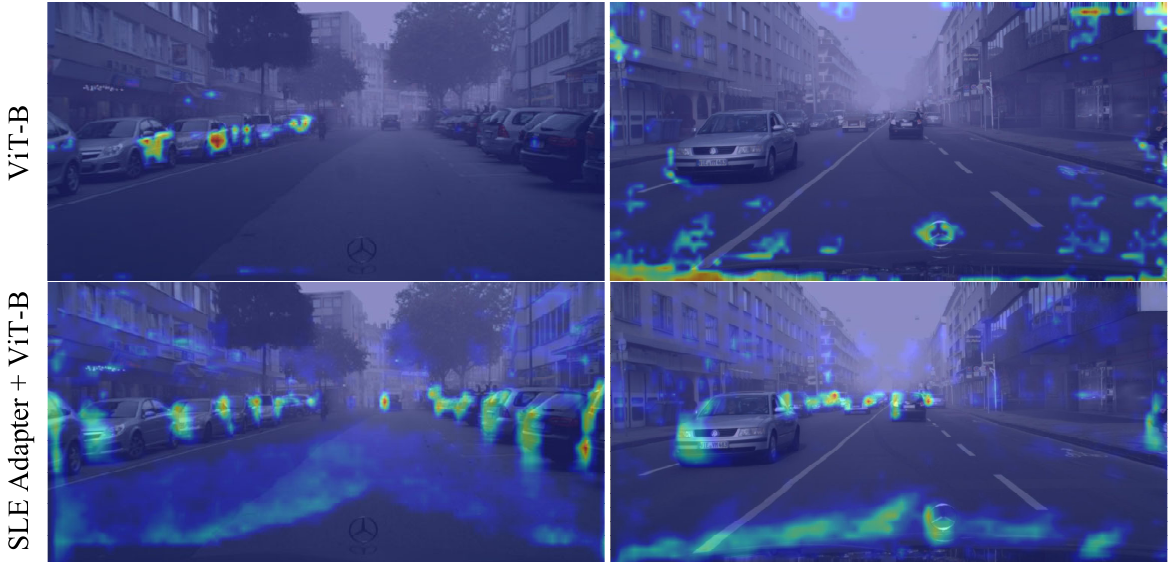}
    \caption{Grad-CAM comparison between raw and SLE-enhanced DINOv2-B.}
    \label{fig:heat_map_sle_comparison}
\end{figure}

SLE Adapter further improves raw-DINOv2 alignment by 0.8/0.9 with VGG4 and 2.1/1.0 with VGG3 on FC/BD.
The larger gain with VGG3 supports the benefit of jointly reformulating the teacher feature and selecting a spatially compatible student feature.
The Grad-CAM comparison in Fig.~\ref{fig:heat_map_sle_comparison} also shows that SLE Adapter emphasizes detection-relevant regions such as cars and pedestrians.
Therefore, although PL is used for the main results, FA provides independent evidence that SLE Adapter improves feature-level transfer rather than only pseudo-label generation.

\begin{table}[tbp]
    \centering
    \begin{tabular}{cccc}
        \toprule
        \textbf{Feature Layer}
        & \textbf{PL}
        & \textbf{FC}
        & \textbf{BD} \\
        \midrule
        VGG4 & --           & 40.1          & 22.3          \\
        VGG4 & $\checkmark$ & 53.6          & 44.3          \\
        VGG3 & --           & 38.7          & 21.7          \\
        VGG3 & $\checkmark$ & \textbf{57.4} & \textbf{47.7} \\
        \bottomrule
    \end{tabular}
    \caption{Pseudo-label-learning ablation on Foggy Cityscapes (FC) and BDD100K Daytime (BD). PL denotes pseudo-label learning. All experiments use pseudo labels generated by SLE-enhanced Dinov2-B.}
    \label{tab:pseudo_label_cross_domain}
\end{table}


\myparagraph{Semantic Compatibility in Pseudo-label Learning.}
Table~\ref{tab:pseudo_label_cross_domain} shows that replacing VGG4 with the stride-16 VGG3 feature improves pseudo-label learning by 3.8 on FC and 3.4 on BD.
VGG3 is weaker without pseudo-labels (38.7 versus 40.1 $\mathrm{AP}_{50}$) but becomes stronger after transfer (57.4 versus 53.6), showing that its advantage comes from more effective knowledge assimilation rather than standalone capacity.
%
These results support student feature selection as an important part of teacher--student semantic compatibility.

\myparagraph{SLE Adapter Components and Efficiency. }
Table~\ref{tab:spm_finetuning_backbone_ablation} isolates the contributions of SLE Adapter.
Injector and Extractor improve DINOv2-B by 3.1; adding a pretrained spatial prior raises the total gain over raw DINOv2-B to 6.7, and fine-tuning the final DINOv2 block adds 1.9 over the adapter without either component.
Combining both choices gives 59.9 $\mathrm{mAP}_{50}$ with DINOv2-B and 61.3 with DINOv2-L.

\begin{figure}[!t]
    \centering
    \includegraphics[width=0.48\textwidth]{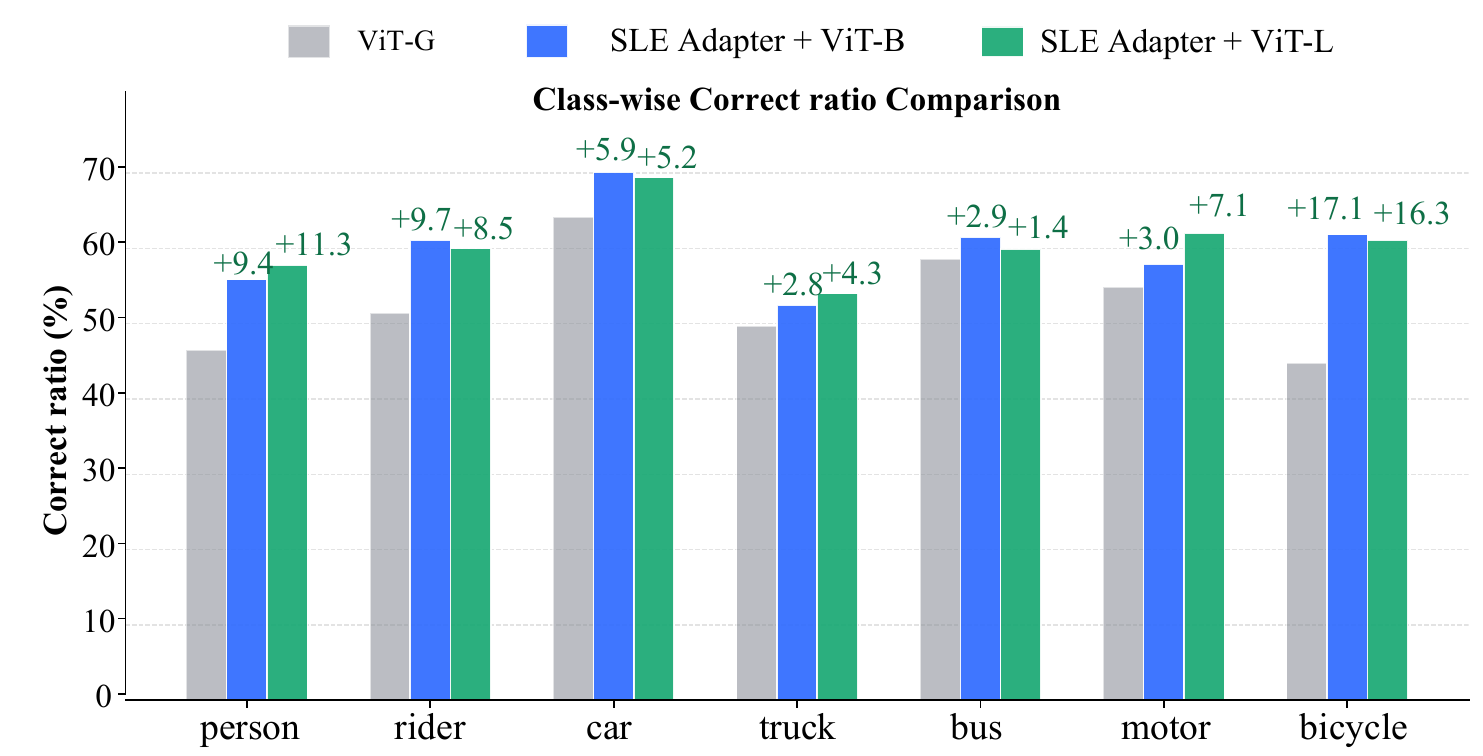}
    \caption{Class-wise correct-pseudo-label ratio for DINOv2 and SLE-enhanced labellers in Foggy Cityscapes.}
    \label{fig:assess}
\end{figure}

SLE Adapter adds about 22M parameters.
As shown in Tab.~\ref{tab:model_size_performance}, SLE with DINOv2-B surpasses DINOv2-G by 2.9 on FC with roughly one-tenth of its model size, while SLE with DINOv2-L improves DINOv2-G by 3.8/1.9 on FC/BD with less than one-third of its size.
Figure~\ref{fig:assess} confirms that these compact teachers also generate more correct pseudo-labels across categories.

\begin{table}[htbp]
    \centering
    \begin{tabular}{cccc}
        \toprule
        \textbf{Backbone}
        & \textbf{Pre-SPM}
        & \textbf{FLB}
        & $\mathbf{mAP}_{50}$ \\
        \midrule
        ViT-B & -- & -- & 51.8 \\
        ViT-L & -- & -- & 54.5 \\
        \midrule
        \multirow{4}{*}{SLE (ViT-B)}
        & --           & --           & 54.9 \\
        & $\checkmark$ & --           & 58.5 \\
        & --           & $\checkmark$ & 56.8 \\
        & $\checkmark$ & $\checkmark$ & 59.9 \\
        \midrule
        SLE (ViT-L)
        & $\checkmark$
        & $\checkmark$
        & \textbf{61.3} \\
        \bottomrule
    \end{tabular}
    \caption{SLE Adapter ablation on Cityscapes $\rightarrow$ Foggy Cityscapes. Pre-SPM and FLB denote a pretrained spatial prior module and fine-tuning the final DINOv2 block, respectively.}
    \label{tab:spm_finetuning_backbone_ablation}
\end{table}

\begin{table}[htbp]
    \centering
    \begin{tabular}{lccc}
        \toprule
        \textbf{Backbone}
        & \textbf{Model Size (M)}
        & \multicolumn{2}{c}{\textbf{$\mathrm{mAP}_{50}$}} \\
        \cmidrule(lr){3-4}
        & & \textbf{FC}
        & \textbf{BD} \\
        \midrule
        ViT-B       & 86    & 51.8          & 43.8            \\
        ViT-L       & 300   & 55.4            & 46.6            \\
        ViT-G       & 1,100 & 57.0          & 50.7          \\
        SLE (ViT-B) & 108    & 59.9          & 49.9          \\
        SLE (ViT-L) & 322    & \textbf{61.3} & \textbf{52.6} \\
        \bottomrule
    \end{tabular}
    \caption{DINOv2 model size and labeller performance on Foggy Cityscapes (FC) and BDD100K Daytime (BD).}
    \label{tab:model_size_performance}
\end{table}

\section{Conclusion}

We identified the spatial-scale discrepancy between VFM teachers and detection students as a source of semantic incompatibility that limits feature alignment and pseudo-label learning.
To address this problem, we proposed SLE-T, which combines SLE Adapter with pseudo-label learning or feature alignment.
SLE Adapter injects local-texture priors into DINOv2 and reformulates its features into dense, student-compatible representations while preserving pretrained semantics.
Experiments on three DAOD benchmarks demonstrate state-of-the-art performance, and the ablations validate the importance of teacher--student semantic compatibility and each SLE component.
Moreover, SLE-enhanced DINOv2-B and DINOv2-L teachers provide competitive or superior target supervision with substantially lower training cost than DINOv2-G.
%
%
These results establish SLE-T as an effective and efficient framework for VFM-based DAOD.

\bibliography{main}
\bibliographystyle{icml2026}

\newpage
\appendix
\onecolumn
\section{Pseudo code of SLE Teacher.}

Algorithm~\ref{alg:sle_teacher} summarizes the Semantic Localization-Enhanced Teacher (SLE-T) framework.
For clarity, $\mathrm{Inj}$ and $\mathrm{Ext}$ denote the deformable-attention operations in the Injector and Extractor, respectively.

\begin{algorithm}[htbp]
\caption{SLE Teacher}
\label{alg:sle_teacher}
\begin{algorithmic}[1]
\REQUIRE $D_s$, $D_t$; transfer route $r\in\{\mathrm{PL},\mathrm{FA}\}$; pretrained DINOv2 $G$ and VGG encoder $E$
\ENSURE Adapted student detector $f_\theta$
\STATE Initialize SPM, $\mathrm{Inj}$, $\mathrm{Ext}$, ConvFFN, and detection head $h$
\STATE Freeze all DINOv2 blocks except the final block $G_{L}$
\FOR{each source mini-batch $(X_s,B_s,Y_s)$}
    \STATE $\mathbf P_{16}\leftarrow
    \mathrm{Flatten}(\mathrm{Conv}_{1\times1}(E^{(16)}(X_s)))$
    \STATE $\mathbf Z_D\leftarrow G_{<L}(X_s)$
    \STATE $\mathbf Z_I\leftarrow
    \mathbf Z_D+\gamma_I\mathrm{Inj}(\mathrm{LN}(\mathbf Z_D),\mathrm{LN}(\mathbf P_{16}))$
    \STATE $\mathbf Z_D^{+}\leftarrow\mathrm{LN}_D(G_L(\mathbf Z_I))$
    \STATE $\mathbf T_0\leftarrow
    \mathbf P_{16}+\mathrm{Ext}(\mathrm{LN}(\mathbf P_{16}),\mathrm{LN}(\mathbf Z_D^{+}))$
    \STATE $\mathbf T\leftarrow
    \mathbf T_0+\mathrm{ConvFFN}(\mathrm{LN}(\mathbf T_0),H_{16},W_{16})$
    \STATE $\mathbf F_T^{\mathrm{SLE}}\leftarrow
    \mathrm{Reshape}(\mathbf T)+
    \mathrm{Interp}(\mathrm{Reshape}(\mathbf Z_D^{+}),H_{16},W_{16})$
    \STATE $\mathcal L_{\mathrm{teacher}}\leftarrow
    \mathcal L_S^{\mathrm{det}}(h(\mathbf F_T^{\mathrm{SLE}}),B_s,Y_s)$
    \STATE Update all unfrozen parameters using
    $\nabla\mathcal L_{\mathrm{teacher}}$
\ENDFOR
\STATE Freeze the trained SLE-enhanced teacher $g_{\mathrm{SLE}}$
\IF{$r=\mathrm{PL}$}
    \STATE Generate $(\tilde B_t,\tilde Y_t)\leftarrow g_{\mathrm{SLE}}(X_t)$
    \FOR{each source--target mini-batch}
        \STATE $\mathcal L\leftarrow
        \mathcal L_S^{\mathrm{det}}+
        \lambda^{\mathrm{unsup}}\mathcal L_T^{\mathrm{det}}$
        \STATE Update $\theta$ using $\nabla_\theta\mathcal L$
    \ENDFOR
\ELSE
    \STATE Set $\mathcal B_A\leftarrow\mathcal B_S$ initially and
    $\mathcal B_S\cup\mathcal B_T$ later
    \FOR{each source and alignment mini-batch}
        \STATE $\mathcal L\leftarrow\mathcal L_S^{\mathrm{det}}+
        \frac{\lambda^{\mathrm{sim}}}{|\mathcal B_A|}
        \sum_{X\in\mathcal B_A}
        \|\phi_{\mathrm{lin}}(\mathbf F_S(X))-
        \mathrm{sg}(\mathbf F_T^{\mathrm{SLE}}(X))\|_1$
        \STATE Update $\{\theta,\phi_{\mathrm{lin}}\}$ using $\nabla\mathcal L$
    \ENDFOR
\ENDIF
\STATE \textbf{return} $f_\theta$
\end{algorithmic}
\end{algorithm}

\section{Instantiate SLE-T with DINOv3}

To evaluate the extensibility of SLE-T beyond DINOv2, we instantiate its SLE-enhanced teacher with DINOv3-B~\cite{simeoni2025dinov3}.
We replace the DINOv2 backbone while retaining the SLE Adapter and training protocol, and evaluate the resulting labeller on all target domains.

\begin{table*}[htbp]
    \centering
    \setlength{\tabcolsep}{7pt}
    \renewcommand{\arraystretch}{1.1}
    \begin{tabular}{lcccccc}
        \toprule
        \multirow{2}{*}{\textbf{Model}}
        & \multirow{2}{*}{\textbf{FC}}
        & \multirow{2}{*}{\textbf{BD}}
        & \multicolumn{4}{c}{\textbf{ACDC}} \\
        \cmidrule(lr){4-7}
        & & &
        \textbf{Fog} &
        \textbf{Night} &
        \textbf{Rain} &
        \textbf{Snow} \\
        \midrule

        DINOv2-G
        & 57.0
        & 50.7
        & 65.9
        & 36.4
        & 45.8
        & 52.0 \\

        SLE-enhanced DINOv2-B
        & 59.9
        & 49.9
        & 72.4
        & 36.5
        & 49.8
        & 60.0 \\

        SLE-enhanced DINOv2-L
        & \textbf{61.3}
        & \textbf{52.6}
        & 70.5
        & \textbf{38.7}
        & 48.2
        & \textbf{60.5} \\

        SLE-enhanced DINOv3-B
        & 59.5
        & 49.7
        & \textbf{73.3}
        & 37.6
        & \textbf{52.5}
        & 60.4 \\

        \bottomrule
    \end{tabular}
    \caption{Labeller performance ($\mathrm{mAP}_{50}$, \%) with different
    backbones on Foggy Cityscapes (FC), BDD100K Daytime (BD), and ACDC.}
    \label{tab:teacher_model_comparison}
\end{table*}

Table~\ref{tab:teacher_model_comparison} shows that DINOv3-B closely matches DINOv2-B on FC and BD, with differences of only 0.4 and 0.2 points, while improving all four ACDC conditions by 0.9--2.7 points.
It also outperforms the larger DINOv2-G on five of the six target domains, including gains of 7.4, 6.7, and 8.4 points on ACDC fog, rain, and snow, respectively.
DINOv2-L remains stronger on FC, BD, night, and snow, whereas DINOv3-B achieves the best fog and rain results.
These results demonstrate that SLE-T generalizes across DINO generations, although its performance remains backbone- and domain-dependent.

\section{Interaction Between Feature Alignment and Pseudo-Label Learning}
Table~\ref{tab:fa_pl_teacher_ablation} shows no synergistic gain from jointly applying feature alignment (FA) and pseudo-label learning (PL).
Compared with PL alone, their combination reduces $\mathrm{mAP}_{50}$ by 0.7 points for DINO Teacher and 1.4 points for SLE-T.
One possible explanation is that FA enforces dense matching to fixed teacher representations, including background and domain-specific cues, whereas PL directly optimizes target localization and classification from selected predictions.
Under a shared weighting and training schedule, these objectives may produce competing gradients that restrict task-specific adaptation.
We therefore use the PL route of SLE-T as the final configuration and leave conflict-aware joint optimization to future work.

\begin{table}[htbp]
    \centering
    \begin{tabular}{lccc}
        \toprule
        \textbf{Method}
        & \textbf{FA}
        & \textbf{Pse}
        & \textbf{$\mathrm{mAP}_{50}$} \\
        \midrule

        Baseline (VGG3)
        & --
        & --
        & 38.7 \\

        Baseline (VGG4)
        & --
        & --
        & 40.1 \\

        \midrule

        \multirow{3}{*}{DINO Teacher}
        & $\checkmark$
        & --
        & 42.6 \\
        
        & --
        & $\checkmark$
        & 53.2 \\
        
        & $\checkmark$
        & $\checkmark$
        & 52.5 \\

        \midrule

        \multirow{3}{*}{SLE Teacher (Ours)}
        & $\checkmark$
        & --
        & 48.9 \\
        
        & --
        & $\checkmark$
        & \textbf{57.4} \\
        
        & $\checkmark$
        & $\checkmark$
        & 56.0 \\

        \bottomrule
    \end{tabular}
    \caption{Ablation study of feature alignment and pseudo-label learning with different teacher models on Foggy Cityscapes. All experiments use pseudo labels generated by SLE-enhanced Dinov2-B.}
    \label{tab:fa_pl_teacher_ablation}
\end{table}

\section{Visualization}

\begin{figure*}[t]
    \centering
    \includegraphics[width=1.0\textwidth]{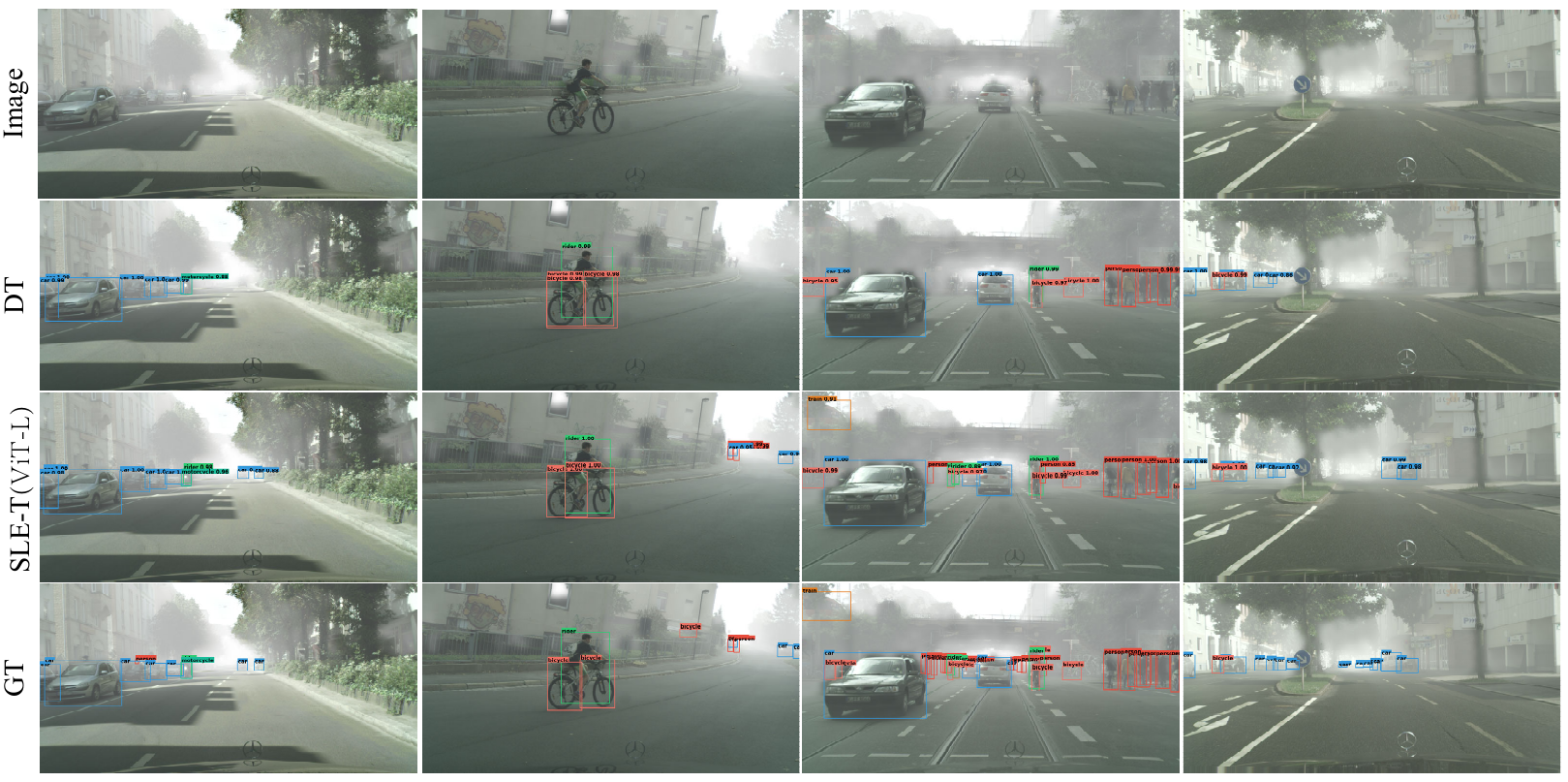}
    \caption{Qualitative detection results on Foggy Cityscapes. From top to bottom, the rows show the input image, predictions from DINO Teacher (DT), predictions from SLE-T instantiated with DINOv2-L, and ground truth (GT).}
    \label{fig:detection_res}
\end{figure*}

Figure~\ref{fig:detection_res} compares student detections produced by DT and SLE-T under dense fog.
Compared with DT, SLE-T recovers more small and distant objects, including car in low-visibility regions, persons in the second scene, and crowded road person in the third scene.
Its predictions also more closely match the GT in the fourth scene, where DT misses several distant vehicles.
These examples suggest that the local cues introduced by SLE Adapter improve the completeness of target pseudo-labels and help the student retain localization-sensitive details, although occasional false positives remain.

\section{Accuracy--Model Size Trade-off}

\begin{figure*}[htbp]
    \centering
    \includegraphics[width=1.0\textwidth]{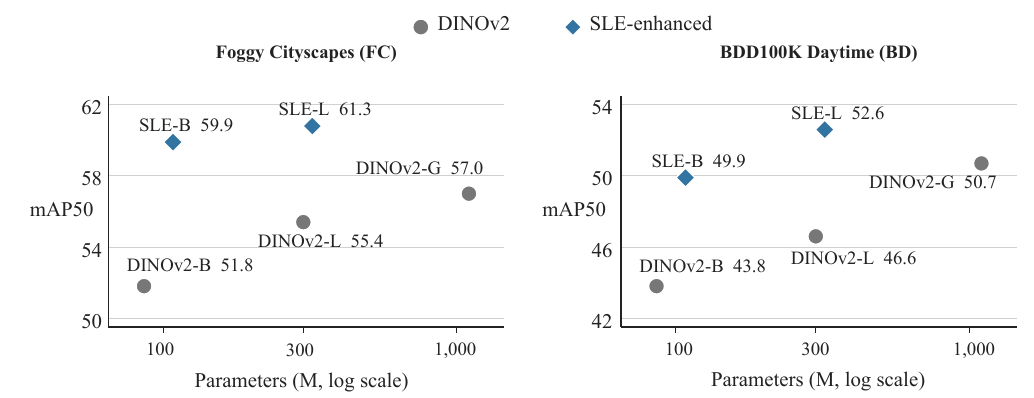}
    \caption{Labeller performance versus model size on Foggy Cityscapes (FC) and BDD100K Daytime (BD). The parameter axis uses a logarithmic scale.}
    \label{fig:model_size_performance}
\end{figure*}

Figure~\ref{fig:model_size_performance} visualizes the accuracy--size trade-off reported in the corresponding table of the main paper.
Adding SLE Adapter shifts DINOv2-B and DINOv2-L substantially upward with only 22M additional parameters, improving FC/BD by 8.1/6.1 and 5.4/6.0 points, respectively.
With 108M parameters, SLE-B outperforms the 1,100M-parameter DINOv2-G by 2.9 points on FC and remains within 0.8 points on BD.
SLE-L further surpasses DINOv2-G by 3.8 and 1.9 points on FC and BD while using less than one-third of its parameters.
The visualization therefore highlights that the gains arise from more effective adaptation rather than simply increasing backbone capacity.

\begin{table*}[htbp]
    \centering
    \setlength{\tabcolsep}{4.5pt}
    \renewcommand{\arraystretch}{1.1}

    \resizebox{\textwidth}{!}{%
    \begin{tabular}{clccccccccc}
        \toprule
        \textbf{\shortstack{Weather\\Condition}}
        & \textbf{Method}
        & \textbf{Person}
        & \textbf{Rider}
        & \textbf{Car}
        & \textbf{Truck}
        & \textbf{Bus}
        & \textbf{Train}
        & \textbf{Motor}
        & \textbf{Bicycle}
        & \textbf{mAP} \\
        \midrule

        \multirow{3}{*}{Fog}
        & DT (ViT-G)
        & 59.9
        & 62.0
        & 78.6
        & 40.0
        & 81.8
        & \textbf{70.1}
        & 36.5
        & 67.5
        & 62.0 \\

        & SLE-T (ViT-B)
        & 64.9
        & \textbf{70.6}
        & 83.5
        & 33.5
        & \textbf{100.0}
        & 63.6
        & 44.2
        & \textbf{69.8}
        & \textbf{66.3} \\

        & SLE-T (ViT-L)
        & \textbf{68.6}
        & 59.0
        & \textbf{84.1}
        & \textbf{41.8}
        & 97.0
        & 63.6
        & \textbf{47.0}
        & 66.8
        & 66.0 \\
        \midrule

        \multirow{3}{*}{Night}
        & DT (ViT-G)
        & 37.1
        & 21.2
        & 58.1
        & 35.6
        & --
        & 40.8
        & \textbf{24.3}
        & \textbf{15.7}
        & 33.3 \\

        & SLE-T (ViT-B)
        & \textbf{40.4}
        & \textbf{35.5}
        & 57.2
        & 30.9
        & --
        & 45.2
        & 24.2
        & 10.8
        & 34.9 \\

        & SLE-T (ViT-L)
        & 39.0
        & 30.6
        & \textbf{59.3}
        & \textbf{37.6}
        & --
        & \textbf{47.7}
        & 19.4
        & 13.0
        & \textbf{35.2} \\
        \midrule

        \multirow{3}{*}{Rain}
        & DT (ViT-G)
        & 43.6
        & 1.7
        & 72.7
        & 28.7
        & 35.7
        & 30.4
        & \textbf{54.2}
        & 0.4
        & 33.4 \\

        & SLE-T (ViT-B)
        & \textbf{50.1}
        & 15.1
        & 78.1
        & 41.6
        & \textbf{39.3}
        & 33.7
        & 50.9
        & 1.6
        & 38.8 \\

        & SLE-T (ViT-L)
        & 48.7
        & \textbf{30.1}
        & \textbf{78.5}
        & \textbf{52.5}
        & 37.0
        & \textbf{38.9}
        & 53.5
        & \textbf{4.2}
        & \textbf{42.9} \\
        \midrule

        \multirow{3}{*}{Snow}
        & DT (ViT-G)
        & 44.6
        & 54.5
        & 69.6
        & 51.3
        & 24.5
        & \textbf{66.8}
        & 49.0
        & 34.5
        & 49.4 \\

        & SLE-T (ViT-B)
        & 54.7
        & 54.5
        & 77.3
        & \textbf{57.6}
        & 25.0
        & 65.4
        & 61.0
        & \textbf{48.7}
        & 55.5 \\

        & SLE-T (ViT-L)
        & \textbf{54.9}
        & \textbf{67.5}
        & \textbf{77.9}
        & 46.5
        & \textbf{30.7}
        & 58.4
        & \textbf{69.6}
        & 40.4
        & \textbf{55.7} \\
        \bottomrule
    \end{tabular}%
    }

    \caption{Class-wise performance of the Cityscapes $\rightarrow$ ACDC results
    reported in the main paper ($\mathrm{AP}_{50}$, \%).}
    \label{tab:c2acdc_classwise}
\end{table*}

\section{Class-wise performance of ACDC Results}

The main paper reports condition-wise $\mathrm{mAP}_{50}$ on ACDC, showing that both SLE-T variants outperform DT under fog, night, rain, and snow.
To identify the categories driving these aggregate improvements, Table~\ref{tab:c2acdc_classwise} provides the corresponding class-wise results.

Both SLE-T variants generally improve person and car detection across adverse conditions, indicating more robust recognition of common road users.
The gains are particularly pronounced for bus and motor under fog and for person, rider, and motor under snow.
Under rain, SLE-T with ViT-L improves rider and truck AP by 28.4 and 23.8 points over DT, respectively, which largely explains its 9.5-point mAP gain.
The two model scales remain complementary: ViT-B is stronger for rider under fog and night and for truck under snow, whereas ViT-L performs better for motor under fog and snow and for rider and truck under rain.
Performance on less frequent classes, such as train, motor, and bicycle, still varies across conditions, suggesting sensitivity to condition-specific sample scarcity.
Overall, the class-wise results explain the condition-level trends in the main paper: the modest night gains are distributed across a few categories, whereas the larger rain and snow improvements arise from substantial gains on several difficult road-user classes.

\end{document}